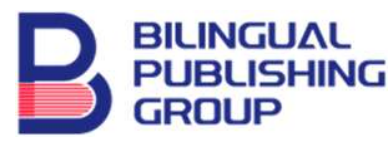

ARTICLE

# Semantic Variational Bayes Based on Semantic Information G Theory for Solving Latent Variables

*Chenguang Lu* [1,2]

[1] *Intelligence Engineering and Mathematics Institute, Liaoning Technical University, Fuxin 123000, China*

[2] *School of Computer Engineering and Applied Mathematics, Changsha University, Changsha 410022, China*

*ORCID: https://orcid.org/0000-0002-8669-0094*

## ABSTRACT

The minimum variational free energy criterion comprises two criteria: the maximum semantic information criterion and the maximum information efficiency criterion, but it does not provide a method for balancing them. The Semantic Information G Theory, the author proposed in his early years, extends the rate-distortion function $R(D)$ to the rate-fidelity function $R(G)$, where $R$ is the minimum mutual information for given semantic mutual information $G$. Semantic Variational Bayes (SVB) is based on the parameter solution of $R(G)$, where the variational and iterative methods originated from Shannon et al.'s research on the rate-distortion function. SVB not only uses likelihood functions but also truth, membership, similarity, distortion, and copula density functions as constraint functions. It explicitly uses the maximum information efficiency ($G/R$) criterion and facilitates the trade-off between maximum semantic information and maximum information efficiency. The computational experiments include 1) using some mixture models as an examples to show that mixture models converges as $G/R$ increases; 2) demonstrating the application of SVB in data compression with a group of error ranges as the constraint; 3) illustrating how the semantic information measure and SVB can be used for maximum entropy control and reinforcement learning in control tasks with given range constraints, providing numerical evidence for balancing control's purposiveness and efficiency. The limitation of SVB is that it does not account for parameter probability distributions. Further research is needed to apply SVB to deep learning.

## 1. Introduction

Machine learning often requires solving the probability distribution $P(y)$ of a latent variable $y$ from observed data $x$ (or the probability distribution $P(x)$), and a group of predictive models or likelihood functions $P(x|y, \theta)$ ($j$ = 1, 2, …). A popular method is the Variational Bayesian (VB) method [1,2]. VB has been successfully applied in various scenarios, such as mixture models [3,4], AutoEncoder [5], active inference with the Minimum Free Energy Principle (FEP) [6], and so on [7]. Bayesian Inference (BI) is an inference method Bayesians use to infer model parameters, including those in the likelihood function $P(x|y, \theta)$ and in parameterized $P(y)$. Unlike in BI, frequentists use Likelihood Inference, which considers only parameters in likelihood functions and does not account for their probability distributions. The author employs frequentist methods and thinks that frequentism should also provide a general method for estimating the latent variable (i.e., the probability distribution $P(y)$).

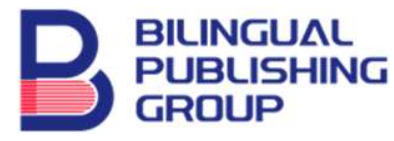

Although the Expectation-Maximization (EM) algorithm used by frequentists can also solve latent variables in mixture models, solving $P(y)$ remains problematic when likelihood functions, i.e., predictive models, are unchanged. This is particularly true when there are fuzzy range constraints rather than likelihood function constraints. For example, in active inference, given the probability distribution $P(x)$ of an uncontrolled state and several targets $y_1$, $y_2$, … (representing fuzzy ranges) as constraints, we need to select an action $a_j$ to achieve a target $y_j$ and optimize $P(a_j)$ for $j$ = 1,2, …. This is a problem that needs to be addressed.

Regularization based on information theory has recently been applied to deep learning [8] and reinforcement learning [9], yielding good results. This method uses the difference between two types of information as the objective function and minimizes it using variational methods to estimate latent variables. The combination of information-theoretic regularization and Variational Bayes has also emerged [10]. This paper shares this goal and hopes to develop a general method that is better understood theoretically. Particularly, providing information efficiency with an upper bound of 1 facilitates a trade-off between maximizing prediction accuracy (or the purposiveness of control) and maximizing information efficiency.

This paper proposes the SVB method. It uses Shannon mutual information minus $s$ times semantic mutual information as the objective function, or say, it uses the maximum information efficiency criterion (compatible with the maximum likelihood criterion) and the maximum entropy principle. Its theoretical basis can be traced to the information rate-distortion theory pioneered by Higgins et al., Shannon and Berger [10–14], including the Arimoto-Blahut algorithm [15].

The author uses the term "Semantic" because SVB is based on a Semantic Information Theory, or G theory [16,17] (G denotes the semantic generalization of the Shannon information Theory). Additionally, SVB uses various learning functions, such as likelihood, truth, membership, similarity, distortion, and copula density functions, as constraints. These functions are related to semantics, according to Davidson's truth-conditional semantics [18]. Although SVB is frequentist, it employs various extended Bayes' formulas and performs tasks similar to those for which VB is used. Hence, the term "Variational Bayes" is still employed.

The G Theory uses various learning functions together with $P(x)$ to express the semantic information measure, i.e., the G measure. It extends the rate-distortion function $R(D)$ to the rate-fidelity function $R(G)$, where $R$ is the minimum Shannon mutual information for given semantic mutual information G. The ratio $G/R$ represents information efficiency. SVB is based on the parameter solution of the $R(G)$ function.

The author proposed the G Theory thirty years ago [19] and has applied it to machine learning in the last decade. Previous papers have discussed methods for solving or optimizing various learning functions from sample distributions and have applied these methods to multi-label classification, maximum mutual information classification, convergence proof for mixture models, Bayesian confirmation, semantic compression, and constraint control [20,21]. Some papers have involved the issue of solving the probability distributions of latent variables.

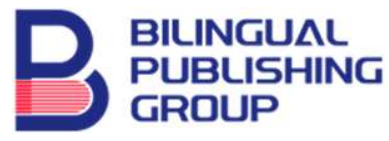

Examining the VB and the Free Energy Principle (FEP), recently, the author found that:

- There are actually two types of variational free energies, $F_1$ and $F_2$. $F_1$ approximates conditional cross-entropy and reflects the inaccuracy of the prediction. Active inference, proposed by Friston et al., commonly uses $F_1$ [6,22]. In contrast, $F_2$ represents the cross-entropy and approximates the source's Shannon entropy. Its negative value is usually interpreted as Evidence Lower Bound. In machine learning, especially in mixture models, $F_2$ is used (see Lemma 2 in Neal and Hinton's article [4]).
- Using $g(y)$ (i.e., $P(y)$ to be optimized) as a variation, the average free energy obtained is $F_1$; while using $g(y|x)$ as a variation, the average free energy obtained is $F_2$ (see Section 3.5). They do not always decrease simultaneously. There are counterexamples in mixture models where $F_2$ decreases while $F_1$ increases during convergence (see Section 4.1).
- Minimizing $F_1$ is equivalent to maximizing semantic mutual information, and minimizing $F_2$ is equivalent to minimizing Shannon mutual information minus semantic mutual information or maximizing information efficiency.

This paper proposes SVB as an alternative to VB for the following reasons:

1) VB is easily misunderstood [21]. For example, it does not clearly distinguish between the two types of variational free energy, $F_1$ and $F_2$ (see Section 3.5 for details), and fails to provide a method for the trade-off between maximizing semantic information (by minimizing $F_1$) and maximizing information efficiency (by minimizing $F_2$).
2) There may be a need to estimate latent variables for given observed data and various constraint functions. We need a more general solution method.

However, SVB also has limitations because it does not account for parameter probability distributions and may not be as useful as VB in some situations.

The primary purposes of this paper are to:

1) Provide a simple and practical frequentist method for solving the probability distribution of latent variables from $P(x)$ and various constraint functions.
2) Provide a method for balancing maximizing semantic information and maximizing information efficiency, with the hope that it can be used in more situations, such as when using beta-VAE [10] for deep learning.
3) Enhance understanding of the G-Theory.

The main contribution of this article is to introduce SVB systematically and to clarify its relationship with VB and FEP.

Since this paper adopts information-theoretic methods, it uses the probability distribution or relative frequency $P(x_i)$ ($i = 1, 2, \ldots m$) of the data instead of the data sequence $x(1), x(2), \ldots, x(N)$. Please note that the expression of the average log-likelihood in this paper is different from that in statistics.

## 2. The Semantic Information G Theory and SVB

This section first introduces the mathematical foundation of G theory: the P-T probabilistic framework, then introduces various learning or constraint functions, semantic information measures, optimization of various learning functions, and the information rate fidelity function $R(G)$ as the basis of SVB.

## 2.1. The P-T Probability Framework and the Semantic Bayes' Formula

The P-T probability framework is used to unify the two types of probability: the set probability defined by Kolmogorov [23] and the probability of elements in sets defined by Mises [24]. The former is logical probability (denoted as $T$), and the latter is the statistical probability (denoted as $P$) adopted by Shannon. The author also follows Zadeh to extend the set probability to the probability of fuzzy sets [25,26].

We define:

- $X$ is a random variable denoting an instance, taking a value $x \in U = \{x_1, x_2, \ldots, m\}$. $Y$ is a random variable denoting a label or hypothesis, taking a value $y \in V = \{y_1, y_2, \ldots, n\}$. The $y_j(x_i)$ is a proposition, $\theta_j$ is a fuzzy subset of $U$, and elements in $\theta_j$ make $y_j$ true. There is $y_j(x) =$ "$x \in \theta_j$". The $\theta_j$ also means a model or a group of model parameters.
- A probability defined with "=", such as $P(y_j) = P(Y = y_j)$, is statistical; a probability defined with "$\in$", such as $P(X \in \theta_j)$, is logical. To distinguish them, we define $T(y_j) \equiv T(\theta_j) \equiv P(X \in \theta_j)$ as the logical probability of $y_j$.
- $T(y_j|x) \equiv T(\theta_j|x) \equiv P(X \in \theta_j|X = x) \in [0, 1]$ is the truth function of $y_j$ and the membership function $m_{\theta j}(x)$ of $\theta_j$, i.e.,

$$T(y_j|x) \equiv T(\theta_j|x) = m_{\theta j}(x). \tag{1}$$

The truth function of $y_j$ indicates its semantics (formal semantics, only related to the extension or denotation of $y_j$). A hypothesis's logical probability may differ from its statistical probability. For example, a tautology's logical probability is 1, whereas its statistical probability is almost 0. We have $P(y_1) + P(y_2) + \ldots + P(y_n) = 1$, but there may be $T(y_1) + T(y_2) + \ldots + T(y_n) > 1$.

According to the above definition, we have:

$$T(y_j) = T(\theta_j) = P(X \in \theta_j) = \sum_i F(x_i) T(\theta_j|x_i). \tag{2}$$

This is the fuzzy event's probability defined by Zadeh [26].

We can put $T(\theta_j|x)$ and $P(x)$ into a Bayes' formula to obtain a likelihood function [17]:

$$P(x|\theta_j) = \frac{T(\theta_j|x)P(x)}{T(\theta_j)}, T(\theta_j) = \sum_i T(\theta_j|x_i)P(x_i). \tag{3}$$

We call Equation (3) the semantic Bayes' formula. Since the maximum of $T(\theta_j|x)$ is 1, from $P(x)$ and $P(x|\theta_j)$, we derive:

$$T(\theta_j|x) = \frac{\frac{P(x|\theta_j)}{P(x)}}{\max_x\left(\frac{P(x|\theta_j)}{P(x)}\right)}. \tag{4}$$

A semantic channel $T(y|x)$ consists of a group of truth functions: $T(\theta_j|x)$, $j = 1, 2, \ldots, n$, as well as a Shannon channel $P(y|x)$ consists of a group of transition probability functions: $P(y_j|x)$, $j = 1, 2, \ldots, n$. When the semantic channel matches the Shannon channel, i.e., $T(\theta_j|x) \propto P(y_j|x)$ or $P(x|\theta_j) = P(x|y_j)$, $j = 1, 2, \ldots, n$, the semantic mutual information reaches its maximum and equals Shannon mutual information.

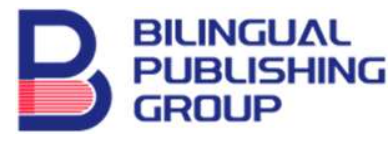

### 2.2. Relationships among Truth, Membership, Similarity, and Distortion Functions

The truth function $T(\theta_j|x)$ of $y_j$ is also the membership function of the fuzzy set $\theta_j$. Assuming that for each $y_j$, there exists an archetype or Platonic ideal $x_j$ such that $T(\theta_j|x_j) = 1$, then the truth value or membership degree $T(\theta_j|x_i)$ is the similarity between $x_i$ and $x_j$. If the domains of $x$ and $y$ are the same, i.e., $U = V$, then $y_j$ becomes an estimate, i.e., $y_j = \hat{x}_j$ = "$x$ is about $x_j$". The similarity function between $x$ and $x_j$, denoted as $S(x, x_j)$, is equivalent to the truth function of $y_j(x)$ and the confusion probability function between them. For instance, consider a Global Positioning System (GPS), the similarity between the indicated location $y_j$ and the actual location $x_i$ is the truth value of $y_j(x_i)$ and their confusion probability.

The truth function and the distortion function can be converted into each other. Let $d(y_j|x_i)$ be the (amount of) distortion of $y_j$ representing $x_i$. We define:

$$T(\theta_j|x_i) = \exp[-d(y_j|x_i)]. \tag{5}$$

Let exp and log be a pair of inverse functions, hence:

$$d(y_j|x_i) = -\log T(\theta_j|x_i). \tag{6}$$

In some cases, it is difficult to directly define the distortion function, so we can first determine the truth function and then use (6) to obtain the distortion function. For example, it is difficult to define the distortion function for the label "elderly" directly. Still, we can use a logistic function as the truth function of "elderly" and then use (6) to obtain the distortion function of "elderly".

Since distortion is generally asymmetric, we use $d(y_j|x_i)$ to represent the distortion of $y_j$ representing $x_i$. For estimation, the distortion function is symmetric. For example, for GPS, the distortion is proportional to the square of distance, and the truth or similarity function is:

$$T(\theta_j|x) = S(x, x_j) = \exp[-d(x, x_j)] = \exp[-\frac{(x-x_j)^2}{2\sigma^2}]. \tag{7}$$

The similarity function is also the observer's discriminant function. A GPS device's accuracy RMS (Root Mean Square of error, denoted as $\sigma$ in the above equation) indicates its resolution or discrimination [16].

Not only does the truth function $T(\theta_j|x)$ represent the semantics of $y_j$. Membership, similarity, and distortion functions all reflect the semantics of labels because a membership function is equivalent to a truth function; the similarity function is a particular case of the truth function; a distortion function can be converted to a truth function.

### 2.3. The Semantic Information Measure

Shannon's mutual information can be expressed as:

$$I(X;Y) = \sum_j \sum_i P(y_j) P(x|y_j) \log \frac{P(x_i|y_j)}{P(x_i)} \tag{8}$$

It represents the average codeword length saved by the probability prediction $P(x|y)$. By replacing $P(x_i|y_j)$ on the right side of "log" with the likelihood function $P(x_i|\theta_j)$ (leaving the left side unchanged), we obtain the semantic mutual information formula:

$$I(X;Y_\theta) = \sum_j \sum_i P(y_j) P(x_i|y_j) \log \frac{P(x_i|\theta_j)}{P(x_i)} = \sum_j \sum_i P(y_j) P(x_i|y_j) \log \frac{T(\theta_j|x_i)}{T(\theta_j)}. \tag{9}$$

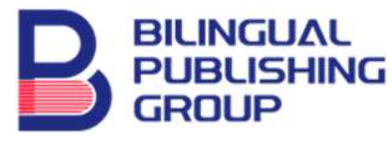

It represents the average codeword length saved by a subjective probability prediction $P(x|\theta_j)$ ($j$ = 1,2,…) according to the semantics of $Y$ [17].

When $Y = y_j$, semantic mutual information becomes semantic Kullback-Leibler (KL) information [17,27]:

$$I(X;\theta_j)=\sum_i P\,(x_i \mid y_j)\log\frac{P(x_i|\theta_j)}{P(x_i)}=\sum_i P\,(x_i \mid y_j)\log\frac{T(\theta_j|x_i)}{T(\theta_j)}. \tag{10}$$

Note that in the above formula, $P(x|y_j)$ is used for averaging and represents the sample distribution. It can be the relative frequency and may not be smooth or continuous. $P(x|\theta_j)$ may differ from $P(x|y_j)$, meaning that information needs factual tests.

Further, if $X = x_i$, the semantic KL information becomes the semantic information conveyed by $y_j$ about $x_i$:

$$I(x_i;\theta_j)=\log\frac{P(x_i|\theta_j)}{P(x_i)}=\log\frac{T(\theta_j|x_i)}{T(\theta_j)}. \tag{11}$$

**Figure 1** illustrates the above formula. It shows that the less the logical probability, the greater the absolute value of the information; the larger the deviation, the less the information; a wrong hypothesis conveys negative information. These conclusions are consistent with Popper's ideas (see Section 10.1-II in [28]). We call $I(x_i;\theta_j)$ and its average the G measure. This measure can represent Popper's verisimilitude (see Section 10.3 in [28]).

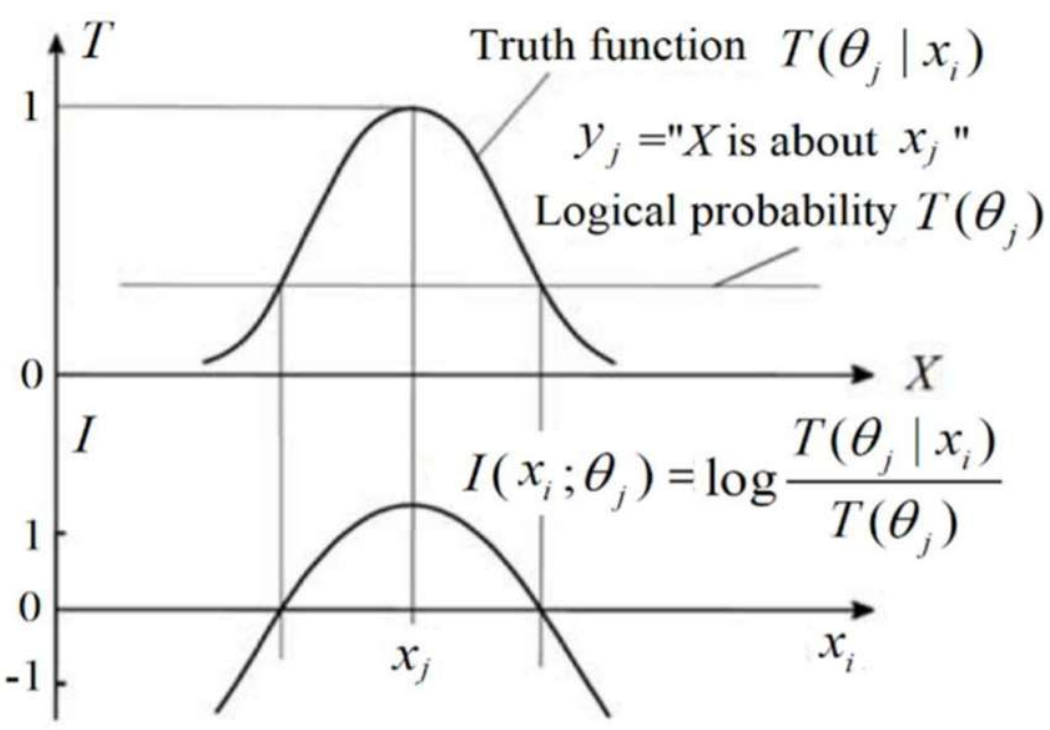

**Figure 1.** Illustrating the amount of semantic information.

Note: The semantic information conveyed by $y_j$ about $x_i$ decreases with the increase in deviation or distortion.

Bring (5) into (9), we obtain

$$I(X;Y_\theta)=-\sum_j P\,(y_j)\log T\,(\theta_j)-\mathrm{E}_{P(x,y)}d(x,y)=H(y_\theta)-\bar{d}. \tag{12}$$

where $H(Y_\theta)$ is the semantic entropy and $\bar{d}$ is the average distortion. $I(X;Y_\theta)$ is like a negative Regularized Least Squares (RLS) measure. Therefore, we can treat the maximum semantic information criterion as a special RLS criterion.

Suppose the truth function in Equation (9) becomes a similarity function. In that case, the semantic mutual information becomes the estimated mutual information, which has been used by deep learning researchers in Information Noise Contrastive Estimation (InfoNCE) [29] and Mutual Information Neural Estimation (MINE) [30].

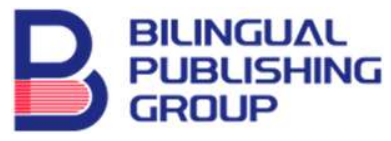

## 2.4. Optimizing Various Learning Functions with Logical Bayes' Inference

The author proposed Logical Bayesian Inference in 2019[17]. Since it is not a Bayesian method, we call it "Logical Bayes' Inference" (LBI) hereafter.

The most often used learning function is the likelihood function $P(x|\theta_j)$. We can use the semantic KL Equation (10) to optimize $P(x|\theta_j)$. Since semantic KL information is equal to the difference between two KL divergences:

$$I(X; \theta_j) = KL(P(x|y_j)||P(x)) - KL(P(x|y_j)||P(x|\theta_j)), \tag{13}$$

it is easy prove that $I(X; \theta_j)$ reaches its maximum when $KL(P(x|y_j)||P(x|\theta_j)) = 0$ or

$$P(x|\theta_j) = P(x|y_j), j = 1, 2, \ldots \tag{14}$$

Sometimes, we wish to use $P(\theta_j|x)$, the parameterized $P(y_j|x)$, as the learning function. Fisher calls $P(\theta_j|x)$ the inverse probability (function) [31]. When $n = 2$, we can use a pair of logistic functions as inverse probability functions. However, it is not easy to construct $P(\theta_j|x)$, $j = 1, 2, \ldots, n$, as $n > 2$, because there is the normalization limit $\sum_j P(\theta_j|x) = 1$ for every $x$. Nevertheless, there is no limit to truth and similarity functions (see Figs. 4a and 4b).

From (4) and (14), we derive the optimized truth function:

$$T^*(\theta_j|x) = \frac{P^*(x|\theta_j)}{P(x)} / \max_x \left(\frac{P^*(x|\theta_j)}{P(x)} = \frac{P(x|y_j)}{P(x)}\right) / \max_x \left(\frac{P(x|y)}{P(x)} = \frac{P(y_j|x)}{\max_x(P(y_j|x))}\right), \tag{15}$$

where "*" means "optimized", $P(x|y_j)$ above is assumed to be a smooth distribution; otherwise, we can only obtain a smooth $T(\theta_j|x)$ by using the following formula:

$$T^*(\theta_j \mid x) = \operatorname{argmax}_{\theta_j} \sum_i P(x_i \mid y_j) \log \frac{T(\theta_j|x_i)}{T(\theta_j)} \tag{16}$$

If we only know $P(y_j|x)$ without knowing $P(x)$, we may let $P(x) = 1/m$ to obtain $T^*(\theta_j|x)$ [17]. We call the above method (of solving truth or similarity functions from sample distributions) LBI.

In the following part of this section, we assume that all likelihood, truth, and similarity functions are optimized, and a sampling distribution is the same as the corresponding probability distribution. We call

$$c(x, y_j) \equiv P(x|\theta_j)/P(x) = P(x, y_j)/[P(x)P(y_j)] \tag{17}$$

the relatedness function. Sklar proposed the copula function [32,33]. The $c(x, y_j)$ is also a two-dimensional copula density function. Since $c(x, y_j)$ is proportional to $P(y_j|x)$, we can replace $P(y_j|x)$ with $c(x, y_j)$ in a Bayes' formula as follows:

$$P(x \mid y_j) = \frac{P(x)c(x,y_j)}{c(y_j)}, c(y_j) = \sum_i P(x_i)c(x_i, y_j), \tag{18}$$

where $P(x)$ may be different from previous $P(x)$. We also call $c(x, y)$ the Bayes' core since $P(x, y) = P(x)c(x, y)P(y)$. When $P(x)$ is unchanged, $P(x|y_j) = P(x_i)c(x, y_j)$. If $P(x)$ is changed, we have to use (18) to obtain new $P(x|y_j)$ and use the Minimum Information Difference (MID) iteration to obtain new $P(y)$ and $c(x, y)$.

The copula density function can also be used as a constraint to find latent variables [34]. For given $P(x)$ and some constraints, if $c(x, y)P(y)$ represents a Shannon channel $P(y|x)$, there should be, for every $j$,

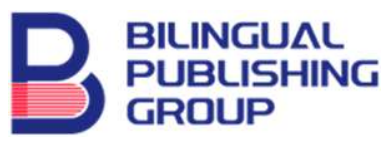

$$\sum_i P(x_i|y_j)=\sum_i P(x_i)c(x_i,y_j)=1 \text{ and } \sum_i P(x_i)c(x_i,y_j)P(y_j)=P(y_j). \tag{19}$$

Otherwise, $P(y)$ and $P(y|x)$ are inappropriate solutions.

If the domains of $x$ and $y$ are the same, then $y_j \equiv \hat{x}_j$ becomes an estimate: "$x$ is about $x_j$." In this case, $T(\theta_j|x_i) = T(\hat{x}_j|x_i) = 1$ as $i = j$, and the truth value becomes the symmetric similarity degree:

$$S(\hat{x}_j|x_i)=\frac{c(x_i,y_j)}{\max_x(c(x,y_j))}=\frac{c(x_i,y_j)}{c(x_j,y_j)}=\frac{c(x_i,y_j)}{c(x_i,y_j)}=S(\hat{x}_i|x_j). \tag{20}$$

However, truth or membership functions may be asymmetric. Suppose all $y$ similar to $y_i$ form a fuzzy subset $\theta_{xi}$ of $V$; we have

$$T(\theta_{xi}|y_j)=\frac{P(x_i|y_j)}{\max_y(P(x_i|y))}\neq\frac{P(y_j|x_i)}{\max_x(P(y_j|x))}=T(\theta_j|x_i). \tag{21}$$

Truth or similarity functions have also been used as learning functions in deep learning [29,30]. For example, Oord et al. [29] introduced InfoNCE and explicitly noted that the learning function is proportional to $m(x, y_j) = P(x|y_j)/P(x)$. The expression in their paper is:

$$f_k(x_{t+k},c_t)\propto P(x_{t+k}|c_t)/P(x_{t+k}), \tag{22}$$

where $c_t$ is the feature vector obtained from previous data, $x_{t+k}$ is the predictive vector, and $f_k(x_{t+k}, c_t)$ is a similarity function (between predicted $x_{t+k}$ and real $x_{t+k}$). The estimated mutual information expressed by a similarity function is a particular case of semantic mutual information.

### 2.5. The Origin of SVB: From R(D) to R(G)

The author uses "information rate fidelity" for two reasons. First, Shannon initially proposed the information rate fidelity criterion [11], and later used minimum distortion as a representation of maximum fidelity. Second, fidelity and verisimilitude are similar.

We change the distortion limit $\bar{d} \leq D$ for $R(D)$ into $I(X; Y_\theta) \geq G$, and then $R(D)$ becomes the information rate-fidelity function $R(G)$, which was proposed in 1993 [19]. In this case, we replace $d_{ij} = d(x_i, y_j)$ with $I_{ij} = I(x_i; \theta_j)$.

In addition to the given $P(x)$ and $G$, there are the following limitations:

$$\sum_i P(x_i|y_j)=1, j=1,2,\ldots,n; \tag{23}$$

$$\sum_j P(y_j)=1. \tag{24}$$

We use the Lagrange multiplier method for Minimum Information Difference (MID):

$$L(P(y|x),P(y))=I(X;Y)-sI(X;Y_\theta)-\mu_j\sum_i P(x_i|y_j)-\alpha\sum_j P(y_j), \tag{25}$$

where $I(X; Y_\theta)$ and $I(X; Y)$ are expressed with $P(y|x)$ and $P(y)$ as:

$$I(X;Y_\theta)=\sum_i P(x_i)\sum_j P(y_j|x_i)\log m_{ij}, m_{ij}=T(\theta_j|x_i)/T(\theta_j)=P(x_i|\theta_j)/P(x_i), \tag{26}$$

$$I(X;Y)=\sum_i P(x_i)\sum_j P(y_j|x_i)[\log[P(y_j|x_i)-\log P(y_j)]. \tag{27}$$

Since $P(y|x)$ and $P(y)$ are interdependent, we alternatively use one as the variation. First, we fix $P(y)$ and let

$$\partial F/\partial P(y_j|x_i)=0, j=1,2,\ldots,n;\ i=1,2,\ldots,m.$$

Hence, we derive

$$P(y_j|x_i) = \frac{P(y_j)m_{ij}^s}{\lambda_i, \lambda_i=\sum_j P(y_j)m_{ij}^s}, i = 1,2,\ldots; j = 1,2,\ldots \tag{28}$$

Note that the above formula is simpler than the solution of $P(y|x)$ in the rate distortion function because there is no exponential operation in $m_{ij}$. We can treat $m^s_{ij}$ as an intensive Bayes' core and consider (28) as a generalized Bayes' formula.

Then, we fix $P(y|x)$ and let

$$\frac{\partial F}{\partial P(y_j)} = 0, j = 1,2,\ldots,n.$$

We derive

$$P(y_j) = \sum_i P(x_i)P(y_j|x_i). \tag{29}$$

Repeating (28) and (29) as we do for the $R(D)$ function [16], we obtain appropriate $P(y)$ and $P(y|x)$. Without this iteration, $m_{ij}{}^s/\lambda_i$, as $c(x_i, y_j)$, is not an appropriate Bayes' core or copula density function, and $P(x|y_j) = P(x)m_{ij}{}^s/\lambda_i$ is not normalized. Equations (28) and (29) implement the MID iteration, also known as the Arimoto-Blahut algorithm, which matches the Shannon channel to the semantic channel.

Finally, we obtain the parameter solution of $R(G)$ (see **Figure 2**):

$$G(s) = \sum_i \sum_j P(x_i)P(y_j|x_i)I_{ij}, R(s) = sG(s) - \sum_i P(x_i)\log\lambda_i. \tag{30}$$

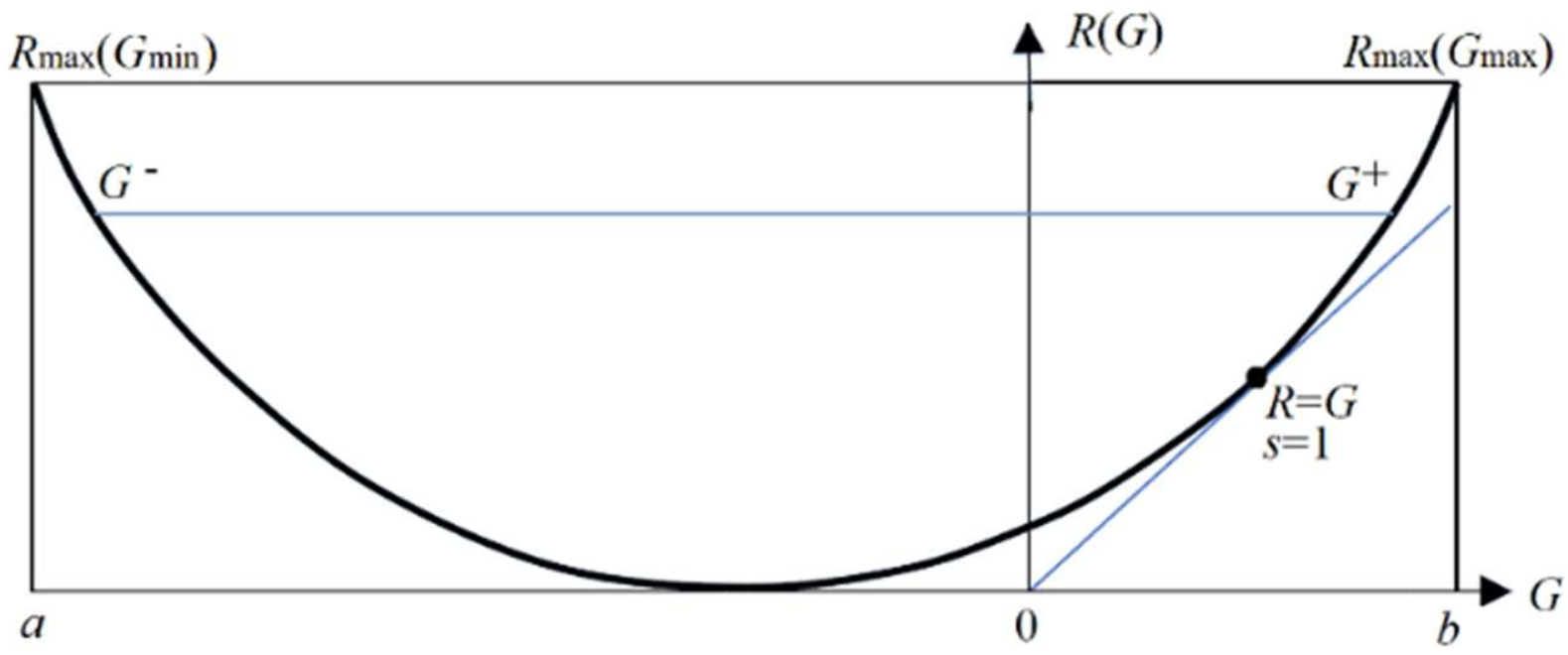

**Figure 2.** The information rate-fidelity function $R(G)$ for binary communication.

Note: Any $R(G)$ function is bowl-like and has a point where $s = 1$ and $R = G$. For given $R$, there are two anti-functions, $G^-(R)$ and $G^+(R)$.

The shape of any $R(G)$ function is a bowl-like curve, which may be asymmetric [16], with the second derivative $\geq 0$. There is $s = \mathrm{d}R/\mathrm{d}G$. When $s = 1$, $R$ equals $G$. $G/R$ indicates the optimized information efficiency. The $R(G)$ function has been applied to image compression based on visual discrimination [16], convergence proofs for maximum mutual information classification and mixture models, semantic compression, and active inference (or constant control) [20].

It is worth noting that, given the semantic channel $T(y|x)$, letting $P(y_j|x) \propto T(\theta_j|x)$ or $P(x|y_j) = P(x|\theta_j)$ does not maximize $G$, but the information efficiency $G/R$. We can increase $s$ to further increase both information with Equation (28). As $s$ -> $\infty$, $P(y_j|x)$ ($j = 1, 2, \ldots, n$) only takes the value 0 or 1, becoming a classification function.

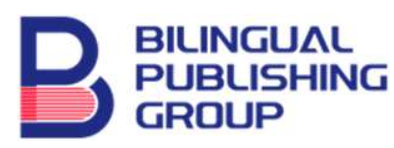

With the Lagrange function in Equation (25), if we fix $P(y|x)$ and $P(y)$ and use $T(\theta_j|x)$ or $P(x|\theta_j)$, $j = 1, 2, \ldots$, as the variation, that is to use LBI to optimize the predictive model. Therefore, we may say that the MID iteration and LBI together form SVB.

We can also replace average distortion $\bar{d}$ with fuzzy entropy $H(Y_\theta|X)$ to obtain the rate-truth function $R(\Theta)$ [20], where $\Theta$ is a group of fuzzy sets as the constraint. $R(G)$ is more suitable than $R(D)$ and $R(\Theta)$ when information is more important than truth. $P(y)$ and $P(y|x)$ for $R(\Theta)$ are different from those for $R(G)$ because the optimization criterion is different. With the maximum truth criterion, $P(y|x)$ becomes

$$P(y_j|x_i) = \frac{P(y_j)[T(\theta_{xi}|y_j)]^s}{\sum_j P(y_j)[T(\theta_{xi}|y_j)]^s}, i = 1,2,\ldots; j = 1,2,\ldots \quad (31)$$

If $T(\theta_{xi}|y) = \exp[-d(x_i|y_j)]$, $R(\Theta)$ becomes $R(D)$. If $T(\theta_j)$ is tiny, $P(y_j)$ for $R(G)$ is larger than that for $R(D)$ and $R(\Theta)$.

## 3. Solving Latent Variables by Using SVB

### 3.1. Approximate Solutions and Optimization Criteria for Latent Variables

From an information-theoretic perspective, we can view the probability distribution $P(x)$ of the observed data as the source, and the required solution $P(y|x)$ as the Shannon channel. The usual constraints are $P(x|\theta_j)$ ($j = 1, 2, \ldots, n$).

Solving $P(y)$ for given $P(x)$ and $P(x|y)$, we can list $m$ equations:

$$P(x_i|y_1)P(y_1) + P(x_i|y_2)P(y_2) + \ldots + P(x_i|y_n)P(y_n) = P(x_i), i = 1, 2, \ldots, m. \quad (32)$$

Adding $P(y_1) + P(y_2) + \ldots + P(y_n) = 1$, we have a total of $m$+1 equations. When $n = m + 1$, we can obtain the exact solution for $P(y)$. When n>m+1, there are multiple solutions. When $n < m + 1$, there is no solution; however, we can obtain an approximate solution that minimizes loss under some criteria.

In the following, we consider only cases where $n < m + 1$ and use the maximum information efficiency criterion to obtain an approximate solution. If a constraint is $T(\theta_j|x)$ instead of $P(x|\theta_j)$, we can use (3) to obtain $P(x|\theta_j) = P(x)T(\theta_j|x)/T(\theta_j)$ as the constraint.

Note that the constraint functions are distributions over $x$, such as $T(\theta_j|x)$ and $s(x, x_j)$ ($j = 1, 2, \ldots, n$), which are not normalized. The functions we need to solve are distributions over $y$, i.e., $P(y|x_i)$ ($i = 1, 2, \ldots, m$). $P(y|x_i)$ must be normalized. Only $P(y|x_i)$ can be placed on the left side of the log for averaging.

The author has considered letting $P(y_j) = T(\theta_j)/\sum_j T(\theta_j)$ and then obtaining $P(y_j|x) = T(\theta_j|x)P(y_j)/T(\theta_j)$. However, this method cannot guarantee $\sum_j P(y_j|x) = 1$ for each $x$.

### 3.2. Proving the EM Algorithm's Convergence

We know $P(x) = \sum_j P(y_j)P(x|y_j)$. Given the sample distribution $P(x)$, we can use $P_\theta(x) = \sum_j P(y_j)P(x|\theta_j)$ to approximate $P(x)$, ensuring that the relative entropy

$$H(P||P_\theta) = \sum_i P(x_i) \log[P(x_i)/P_\theta(x_i)] \quad (33)$$

approaches zero. $P(y)$ is the probability distribution of the latent variable $y$.

The EM algorithm initially sets $P(x|\theta_j)$ and $P(y_j)$, $j = 1, 2, \ldots, n$. Then, in the E-step, we obtain:

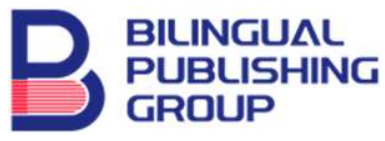

$$P(y_j|x) = P(y_j)P(x|\theta_j)/P_\theta(x), P_\theta(x) = \sum_k P(y_k)P(x|\theta_k). \tag{34}$$

In the M-step, we maximize the complete-data log-likelihood. This step includes the M1-step:

$$P^{+1}(y_j) = \sum_i P(x_i)P(y_j|x_i) \tag{35}$$

and the M2-step：

$$P(x|\theta_j^{+1}) = P(x)P(y_j|x)/P^{+1}(y_j) = P(x)\frac{P(x|\theta_j)}{P_\theta(x)}\frac{P(y_j)}{P^{+1}(y_j)}, \tag{36}$$

which optimizes a group of likelihood functions. For Gaussian mixture models, we use the right-side expectation and standard deviation as those of the left side.

This iterative process continues until the mixture model converges. According to the derivation of the information rate-fidelity function, the E-step and M1-step minimize the information difference $R - G$. According to (14), the M2-step maximizes $G$. Therefore, the EM algorithm uses the maximum information efficiency criterion.

The author tried to improve the EM algorithm to the EnM algorithm [17]. The EM algorithm includes the E-step of the EM algorithm, the n-step, and the M2-step of the EM algorithm. The M2-step is also used as the new M-step by Neal and Hinton [4]. The n-step repeats the E-step and the M1-step $n$ times, such that $P^{+1}(y) \approx P(y)$. Clearly, the EnM algorithm also uses the maximum information efficiency criterion. The n-step can accelerate the convergence of $P(y)$. However, since the E-step takes longer than the M2-step, the EnM algorithm may be uneconomical.

The following equations can be used to prove the EM algorithm's convergence. After the E-step, Shannon's mutual information $I(X; Y)$ becomes

$$R = \sum_i \sum_j P(x_i)\frac{P(x_i|\theta_j)}{P_\theta(x_i)}P(y_j)\log\frac{P(y_j|x_i)}{P^{+1}(y_j)}. \tag{37}$$

We define

$$R'' = \sum_i \sum_j P(x_i)\frac{P(x_i|\theta_j)}{P_\theta(x_i)}P(y_j)\log\frac{P(x_i|\theta_j)}{P_\theta(x_i)}. \tag{38}$$

Then we can derive that after the E-step, there is [17]:

$$H(P||P_\theta) = R'' - G = R - G + H(P_Y{}^{+1}||P_Y), \tag{39}$$

where $H(P||P_\theta)$ is the relative entropy or the KL divergence about $X$; another relative entropy is:

$$H(P_Y{}^{+1}||P_Y) = \sum_j P^{+1}(y_j)\log[P^{+1}(y_j)/P(y_j)]. \tag{40}$$

The latter is close to zero after the $n$-step or the M1-step.

We can use (39) to prove mixture models' convergence because the M2-step matches the semantic channel to the Shannon channel to maximize $G$. And, the E-step and the M1-step match the Shannon channel to the semantic channel and minimize $R–G$ and $H(P_Y{}^{+1}||P_Y)$, making $H(P||P_\theta)$ approach zero.

### 3.3. The Channel Mixture Model and Data Compression

The likelihood functions or components $P(x|\theta_j)$ ($j$ = 1, 2, …, $n$) of a mixture model can also be obtained from the semantic channel $T(\theta_j|x)$ ($j$ = 1, 2, …, $n$) and $P(x)$. Given $P(x)$ and the semantic channel, finding the minimum Shannon mutual information is a data-compression problem. If we simultaneously maximize the semantic mutual information by adjusting the parameters of the semantic channel, we obtain the channel mixture model.

In the EM algorithm, $I_{ij}$ = log[$P(x|\theta_j)/P(x_i)$]. If likelihood functions become truth functions, we have $I_{ij}$ = log[$T(\theta_j|x_i)/T(\theta_j)$]. The E-step becomes:

$$P(y_j|x_i) = \frac{P(y_j)P(x_i|\theta_j)}{\sum_k P(y_k)P(x_i|\theta_k)} = \frac{P(y_j)T(\theta_j|x_i)/T(\theta_j)}{\sum_k P(y_k)T(\theta_k|x_i)/T(\theta_k)}. \quad (41)$$

Assuming $T(\theta_j|x)$ follows a Gaussian distribution, we can use the mean and standard deviation of $P(y_j|x)$ as those of $T(\theta_j|x)$ when optimizing $T(\theta_j|x)$. Using (41) and (29), we can determine the probability distribution $P(y)$ of the latent variable $y$. Adding $s$ to the constraint, as in (28), yields the parameter solution of the $R(G)$ function for data compression.

The truth function $T(\theta_j|x)$ of $y_j$ represents the fuzzy range to which $x$ belongs. The distortion constraint is a particular case of the range constraint. If a distortion function is given instead of a truth function, we may use $T(\theta_j|x) = \exp[-d(x, y_j)]$ as the truth function, and the rest remains the same. If we use the minimum distortion criterion instead of the maximum semantic information criterion, Equation (41) becomes:

$$P(y_j|x_i) = \frac{P(y_j)\exp[-d(x_i,y_j)]}{\sum_k P(y_k)\exp[-d(x_i,y_j)]}. \quad (42)$$

If we add $s$ to (42), we can use the MID iteration to obtain the parameter solution of the $R(\Theta)$ or $R(D)$ function.

### 3.4. Optimizing Latent Variables in Constraint Control and Reinforcement Learning

The information max-min method [8,9] has already been applied in constraint control and reinforcement learning. SVB should be able to provide a more concise approach.

The G measure can also measure goal-oriented information, indicating purposiveness, in controlling random events. An uncertain control objective can be represented by a truth or membership function. For example, there are the following objectives:

- "Crop yields should be close to or exceed 7,500 kg/hectare."
- "Workers' wages are preferably above $5,000."
- "People's life expectancy had better exceed 80 years."
- "The error of train arrival time is preferably within 1 minute."

To measure the goal-oriented information, we can use the semantic KL formula:

$$I(X; a_j/\theta_j) = \sum_i P(x_i|a_j)\log\frac{T(\theta_j|x_i)}{T(\theta_j)}, \quad (43)$$

where $\theta_j$ is a fuzzy set representing a fuzzy range as the control objective; $a_j$ denotes the action taken for the corresponding control task $y_j$. If there are several control objectives $y_1$, $y_2$,…, we can use the semantic mutual information formula to express goal-oriented information:

$$I(X; A/\theta) = \sum_j P(a_j)\sum_i P(x_i|a_j)\log\frac{T(\theta_j|x_i)}{T(\theta_j)}, \quad (44)$$

where $A$ is a random variable taking a value $a_1$, $a_2$, …, or $a_n$, and $P(a)$ is the probability distribution of the latent variable $a$. We need to optimize $P(a)$ using the maximum information efficiency criterion. Unlike the mixture model, here, we do not need to optimize $P(x|\theta_j)$ to make $P_\theta(x)$ close to $P(x)$.

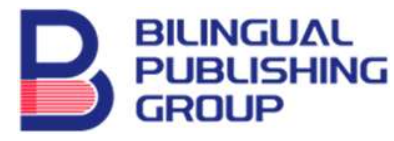

The formula for goal-oriented control information (i.e., the semantic information of imperative sentences) is the same as for the semantic information of descriptive (or predictive) sentences or labels, but their optimization methods differ. For descriptive sentences, the fact $P(x|y_j)$ is fixed, and we hope that the semantic probability prediction $P(x|\theta_j)$ accords with the fact. For imperative sentences, we hope that the fact $P(x|a_j)$ accords with the purpose $P(x|\theta_j)$.

Semantic mutual information $G = I(X; A/\theta)$ represents the purposiveness of control, whereas Shannon's mutual information $R = I(X; A)$ indicates the complexity of control. For multiple-objective tasks, we need to minimize $R = I(X; A)$ subject to the given $G = I(X; A/\theta)$ and the normalization constraints (see the Lagrangian function in Equation (25)). When the actual distribution $P(x|a_j)$ approaches the constrained distribution $P(x|\theta_j)$, information efficiency (not information) reaches its maximum value of 1.

Using (28) and (29), we obtain:

$$P(a_j|x) = \frac{P(a_j)m_{ij}^s}{\lambda_i}, \lambda_i = \sum_j P(y_j)m_{ij}^s, \tag{45}$$

$$P(a_j) = \sum_i P(x_i)P(a_j|x_i). \tag{46}$$

Repeating the above two equations until $P(a)$ remains unchanged ensures that $\frac{m_{ij}^s}{\lambda_i}$ becomes an appropriate Bayes' core. Then we have:

$$P(x_i|a_j) = \frac{P(x_i)m_{ij}^s}{\lambda_i}. \tag{47}$$

We can increase the goal-oriented information by increasing $s$, obtaining the latent variable's distribution $P(a)$ that changes with $s$.

Since the optimized $P(x|a_j)$ is a function of $\theta_j$ and $s$, we write $P^*(x|a_j) = P(x|\theta_j, s)$. It is worth noting that the many distributions of $P(x|a_j)$ satisfy the constraint and maximize $I(X; a_j/\theta_j)$, but only $P^*(x|a_j)$ minimizes $I(X; a_j)$. Assuming that the actual control result is a Gaussian likelihood function $P(x|\beta_j)$, we can replace $P(x|\theta_j, s)$ with $P(x|\beta_j)$, that is, use the expectation and standard deviation of $P(x|\theta_j, s)$ as those of $P(x|\beta_j)$, and obtain an approximate optimal control result.

Assuming we know the effects of actions in reinforcement learning without needing to learn $P(x|\beta_j)$, the above method can also optimize reinforcement learning.

### 3.5 Two Types of Variational Free Energy for Maximizing Semantic Information and Maximizing Information Efficiency

A popular definition of variational free energy is:

$$F = \boldsymbol{E}_{g(y)} \log \frac{g(y)}{P(x,y|\theta)} = \sum_Y g(y) \log \frac{g(y)}{P(x|y,\theta)P(y)} \tag{48}$$

According to Neal and Hiton's article [4], $g(y)$ is a shorthand for $g(y|x)$. Accordingly, the average free energy is:

$$F_2 = \sum_i \sum_j P(x_i) g(y_j|x_i) \log \frac{g(y_j|x_i)}{P(y_j)} - \sum_i \sum_j P(x_i) g(y_j|x_i) \log P\,(x_i|y_j, \theta). \tag{49}$$

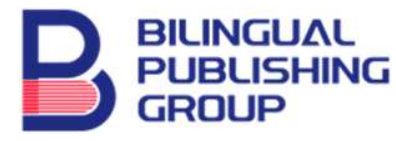

where $g(y|x)$ is the above $P^{+1}(y|x)$, and $P(x|y, \theta)$ is the above $P(x|\theta_j)$. After minimizing $F_2$, $g(y) \approx P(y)$, so we have:

$$F_2 \approx I(X; Y) + H(X/Y_\theta) = R - G + H(X). \tag{50}$$

In active inference, most authors use the following average free energy formula [6,23], where $g(y)$ on the right-hand side of the log does not recover to $g(y|x)$, so they obtain

$$F_1 = \sum_i \sum_j P(x_i) g(y_j|x_i) \log \frac{g(y_j)}{P(x_i|y_j,\theta)} \approx H(X|Y_\theta) = H(X) - G. \tag{51}$$

It approximates the conditional cross-entropy $H(X|Y_\theta)$. The author also previously understood the average variational free energy as $F_1$ [20,21], as in Friston et al., but this understanding is incomplete.

Although Sengupta et al. [35] also affirm that minimum free energy corresponds to maximum information efficiency, they do not specify which information corresponds to cost or utility, or how to calculate information efficiency.

Now we can see that minimizing $F_1$ is equivalent to maximizing $G$; minimizing $F_2$ is equivalent to minimizing the information difference $R - G$ or maximizing the information efficiency $G/R$. Therefore, SVB and VB use similar objective functions, but VB cannot be used for a trade-off between maximizing $G$ and maximizing $G/R$.

Many people overlook a problem: there exists a simple solution for the Shannon channel $g(y|x)$ that minimizes $F_1$, namely $g^*(y|x)$ is a Dirac function concentrated at $x$ and $y$ where the likelihood function is maximum. If only $F_1$ is used to optimize variational inference, this method may degenerate into error control.

In SVB, the objective function to be minimized is

$$f = R - sG. \tag{52}$$

Adjusting $s$ can achieve a trade-off between maximum utility information $G$ and minimum cost information $R$, and $1/s$ is the marginal utility information rate.

In the beta-VAE method proposed by Higgins et al. [10], the objective function is $F_1 - \beta R$. Although adjusting $\beta$ can also achieve a trade-off between maximum semantic information and maximum information efficiency, it is difficult to quantify information efficiency, as its upper bound is 1.

## 4. Experimental Results

### 4.1. A Typical Example where $F_2$ Decreases while $F_1$ Increases

**Table 1** and **Figure 3** illustrate a binary mixture model where the initial two standard deviations are smaller than those of the true model. During the iteration process, $F_1$ increased, and $Q$ decreased.

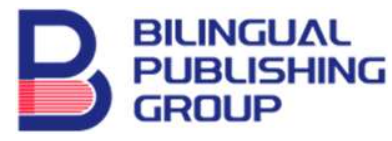

**Table 1.** A mixture model whose $F_1$ increased and $Q$ decreased in the convergent process.

| | The True Model's Parameters | | | Initial Parameters | | |
|---|---|---|---|---|---|---|
| | $c^*$ | $\sigma^*$ | $P^*(Y)$ | $c$ | $\sigma$ | $P(Y)$ |
| $y_1$ | 40 | 15 | 0.5 | 40 | 5 | 0.5 |
| $y_2$ | 75 | 15 | 0.5 | 40 | 5 | 0.5 |

Note:

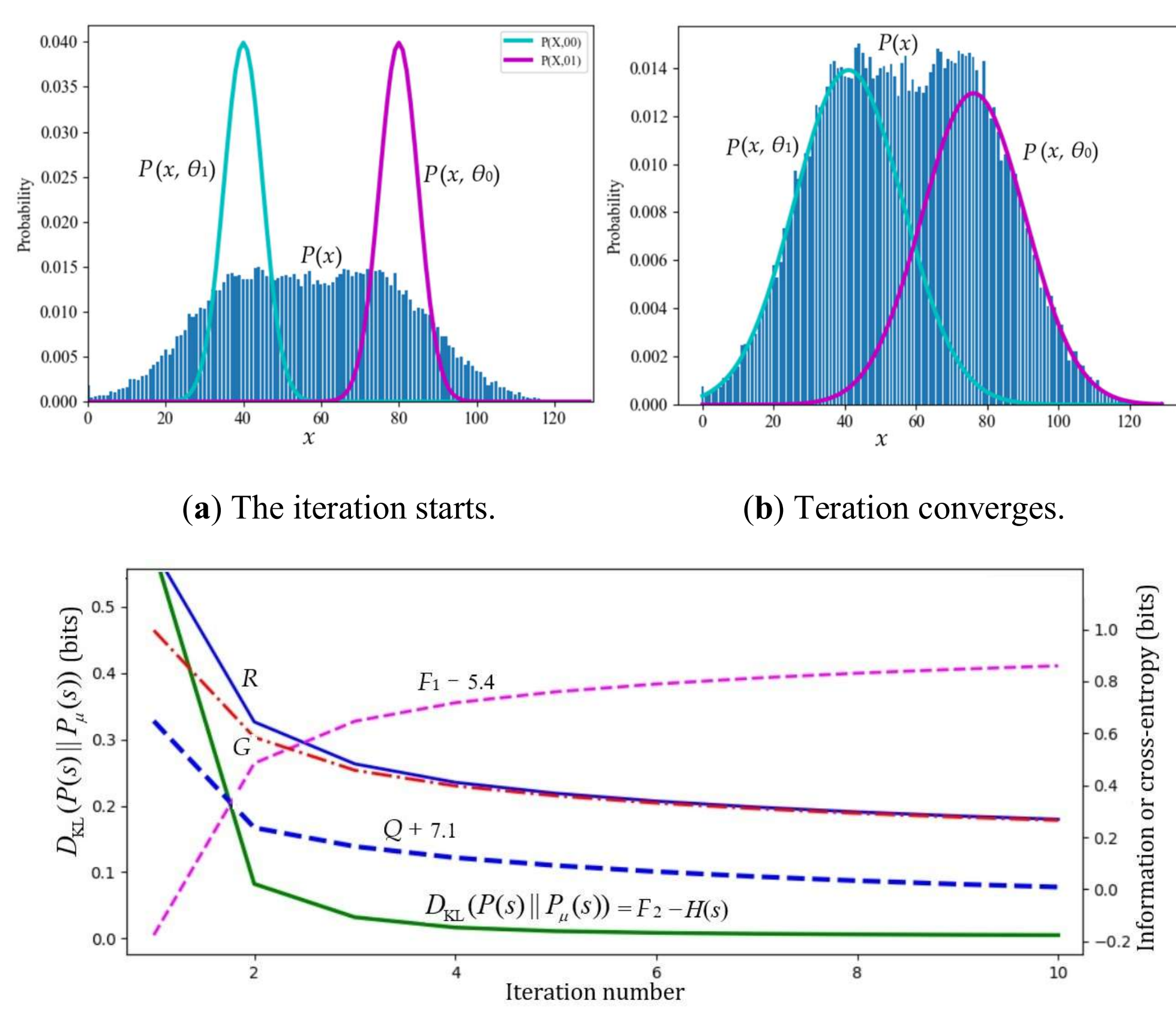

(**a**) The iteration starts.

(**b**) Teration converges.

(**c**) $F_2$, $R - G$, $F_1$, and $Q$ change in the iteration process.

**Figure 3.** During the convergence of the mixed model, $R$–$G$ and $F_2$ decreased, while $F_1$ increased, indicating that $F_1$ and $F_2$ do not always decrease simultaneously.

It is unexpected to many that $F_1$ increased and $Q$ decreased, which contradicts a well-known convergence proof [36,37].

### 4.2. Data Compression According to the Maximum Information Efficiency Criterion

It was supposed that four labels $y_1$ = "non-adult", $y_2$ = "young person", $y_3$ = "adult", and $y_4$ = "elderly" had truth functions as shown in **Figure 4a**. The task was to use the maximum information efficiency criterion to obtain the Shannon channel $P(y|x)$ for a given $P(x)$ with the four truth functions as constraints ($s$ = 1).

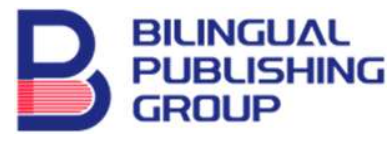

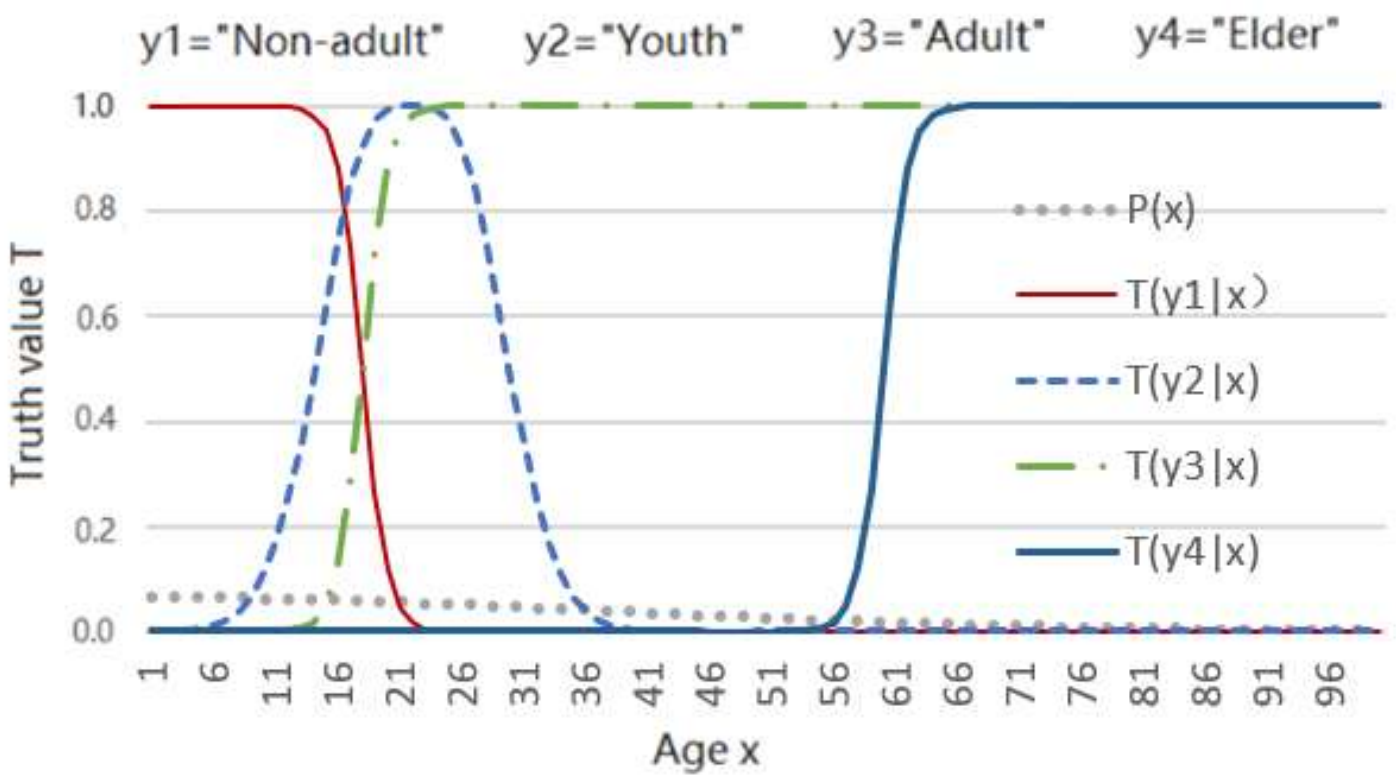

(**a**) The truth functions of four labels over ages.

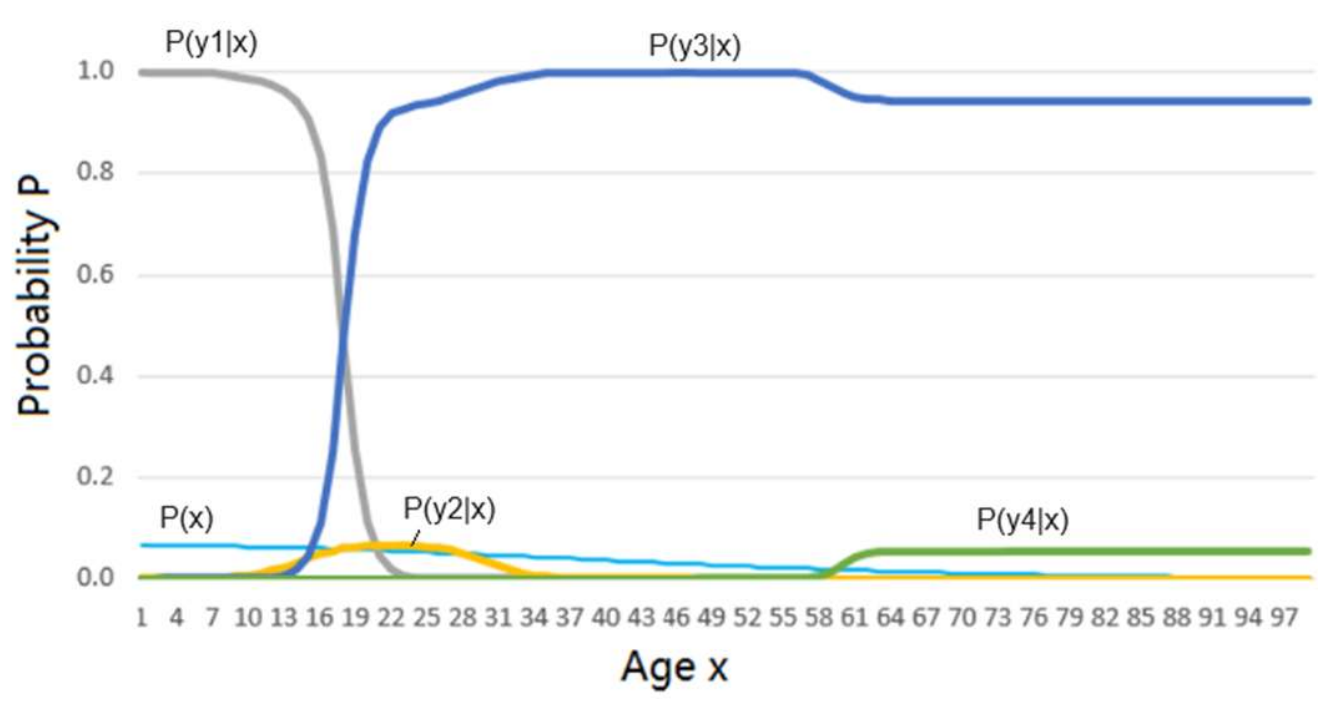

(**b**) The convergent Shannon channel $P(y|x)$.

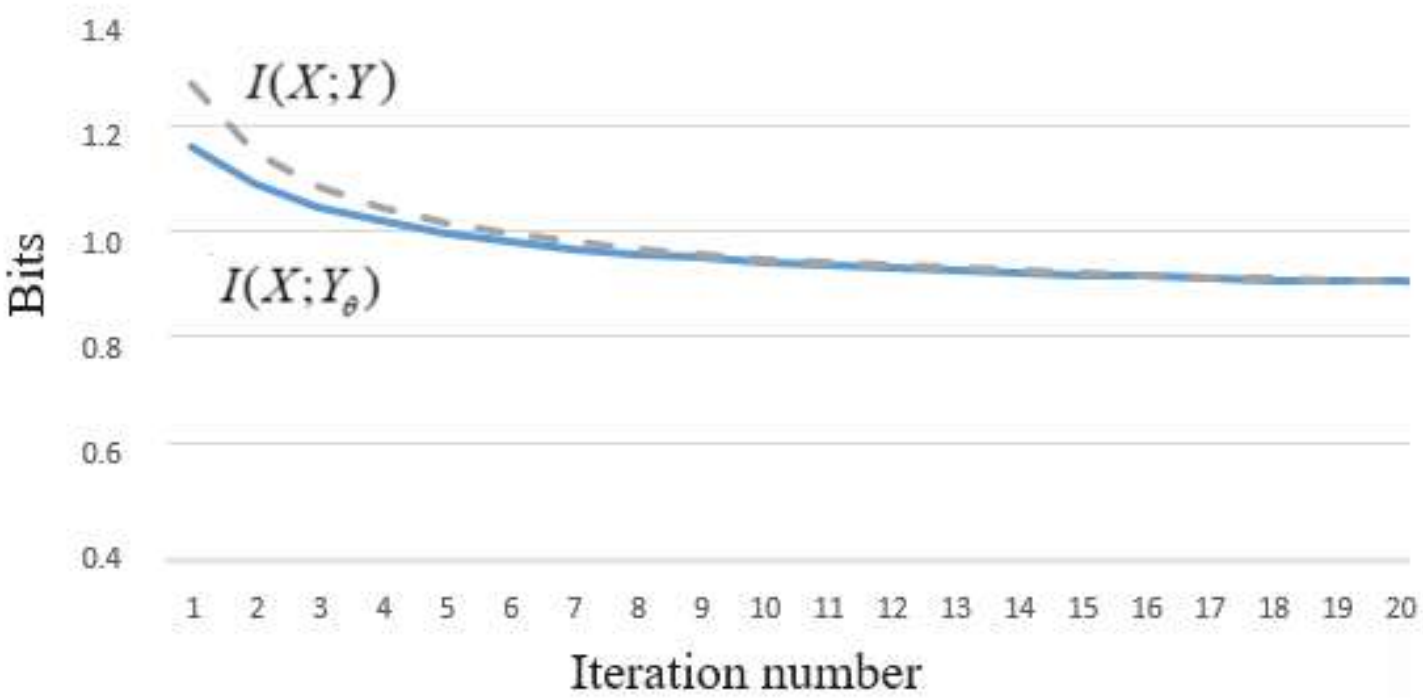

(**c**) The changes of $G$=$I$($X$; $Y_\theta$) and $R$ = $I$($X$; $Y$) during the iterative process.

**Figure 4.** Finding $P(y|x)$ with MID $R - G$ for given constraint ranges.

**Figure 4b** shows that $P(y_j|x)$ and $T(\theta_j|x)$ cover the same areas for $j$ = 1,2,3,4. However, their maximum values differ. **Figure 4c** shows that $I(X; Y)$ minus $I(X;Y_\theta)$ gradually converges over iterations. $P(y_1)$ and $P(y_3)$ are larger than $P(y_2)$ and $P(y_4)$ for less Shannon mutual information $R$. The minimum $R$ is 0.883 bits.

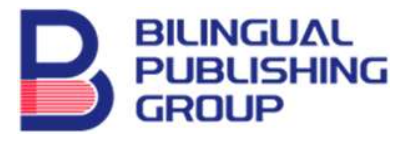

### 4.3. Multi-Objective Control: Optimizing Control Proportions According to Difficulties

**Figure 5** shows a two-objective control task, with objectives represented by the truth functions $T(\theta_0|x)$ and $T(\theta_1|x)$. We can imagine these as two pastures with fuzzy boundaries where we need to herd sheep. Without control, the density distribution of the sheep is $P(x)$. We need to solve an appropriate distribution $P(a)$.

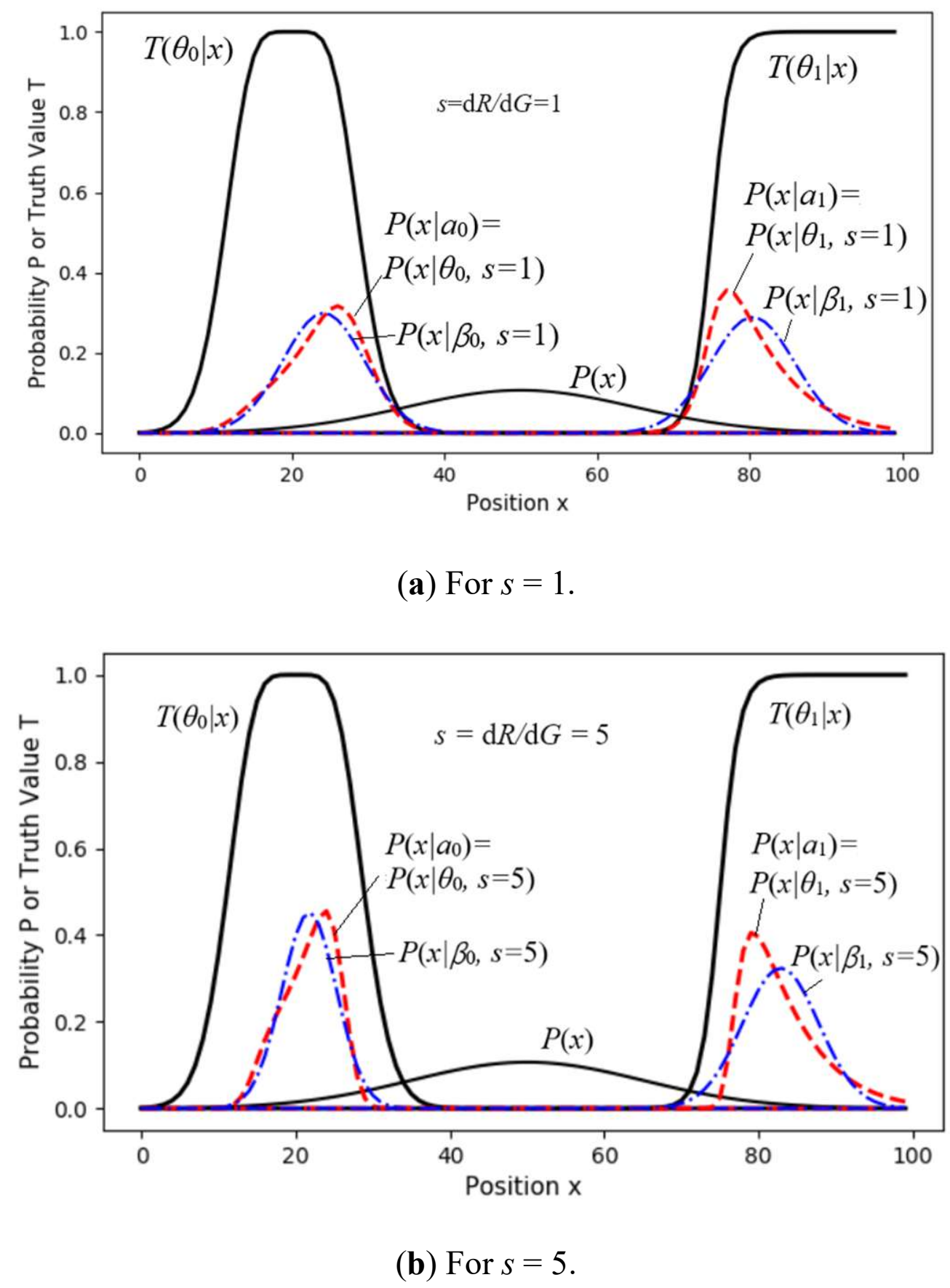

(**a**) For $s$ = 1.

(**b**) For $s$ = 5.

**Figure 5**. A two-objective control task. Dashed lines show $P(x|a_j) = P(x|\theta_j, s)$ ($j$ = 0, 1), and dash-dotted lines representine $P(x|\beta_j, s)$ ($j$ = 0,1). $P(x|\beta_j, s)$ is a normal distribution produced by action $a_j$.

In **Figure 5**, $P(x)$ is a normal distribution with a mean $\mu$=50 and a standard deviation $\sigma$ = 15. The two truth functions representing the two objectives are:

$$T(\theta_0|x) = 1- [1 - \exp(-(x-20)^2/(2\times 25))]^3, \quad (53)$$

$$T(\theta_1|x) = 1/[1+\exp[-0.8(x - c)]. \quad (54)$$

For different $s$, the author set the initial proportions to $P(a_0) = P(a_1) = 0.5$. Then, he used Equations (46) and (47) for the MID iteration to obtain reasonable $P(a_j|x)$ ($j$ = 0,1). Then he obtained $P(x|a_j) = P(x|\theta_j,s)$ using (48). Finally, he obtained $G(s)$, $R(s)$, and $R(G)$ by using (30).

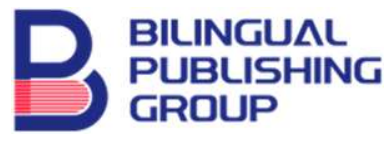

**Table 2** shows the changes of $P(a)$ with $s$ in (46) and $c$ in (50). **Figure 6** shows the $R(G)$ function. **Table 2** indicates that $G$ and $R$ increase and $G/R$ decreases with $s$ increasing; $P(a_1)$ decreases with $c$ increasing. The variation in $P(a)$ shows that the iterative algorithm can reduce the control ratios of difficult tasks.

**Table 2.** $R(G)$, $P(a)$, and $G/R$ change with $s$ and $c$.

| *s* | *c* | $P(a_0)$ | $P(a_1)$ | $G$ (bits) | $R$(bits) | $G/R$ |
|---|---|---|---|---|---|---|
| 1 | 75 | 0.535 | 0.465 | 3.43 | 3.43 | 1 |
| 1 | 80 | 0.579 | 0.421 | 3.80 | 3.80 | 1 |
| 5 | 75 | 0.540 | 0.460 | 3.89 | 4.29 | 0.907 |
| 5 | 80 | 0.592 | 0.408 | 4.28 | 4.71 | 0.909 |
| 40 | 75 | 0.540 | 0.460 | 3.95 | 5.01 | 0.803 |
| 40 | 80 | 0.592 | 0.408 | 4.33 | 5.34 | 0.811 |

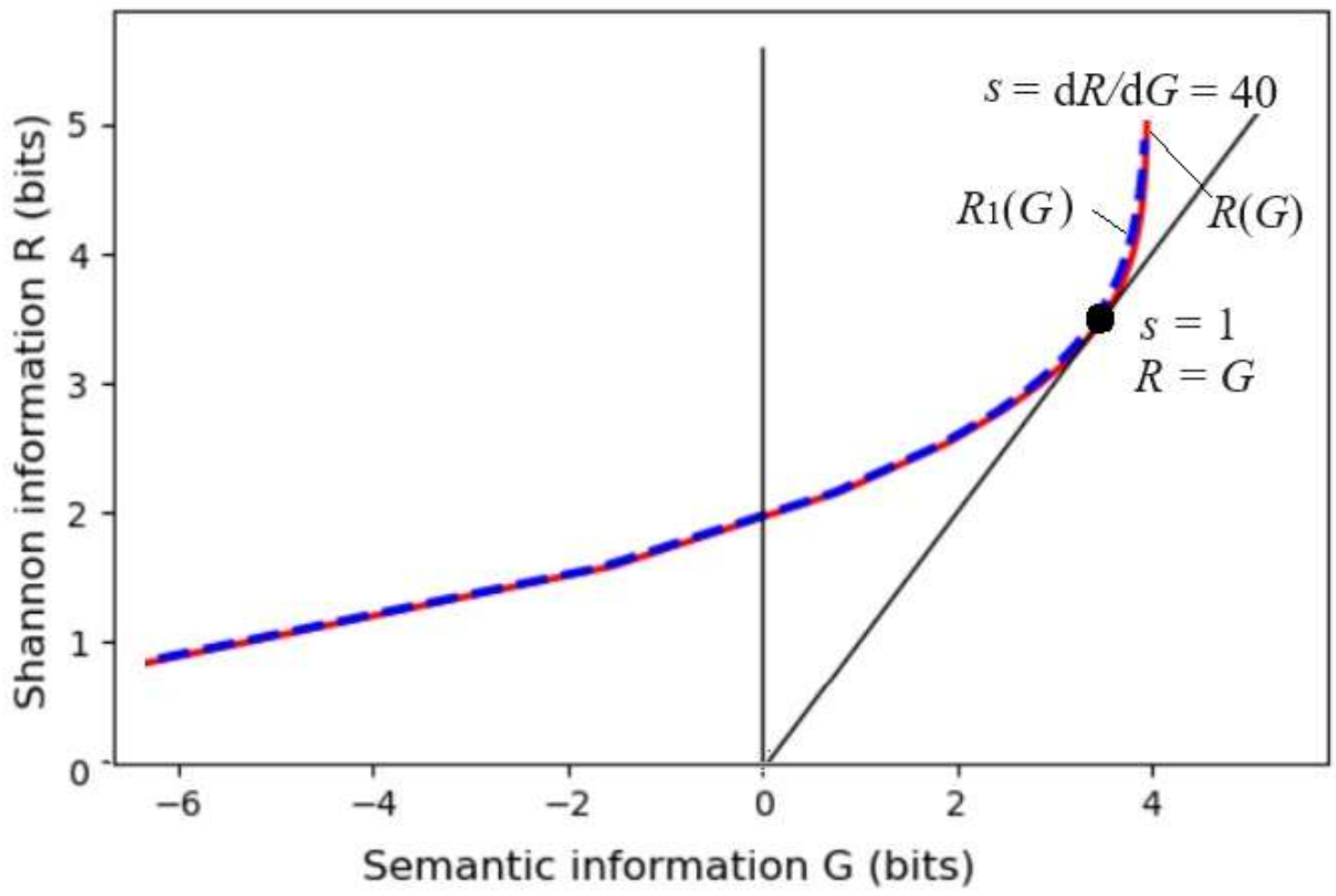

**Figure 6.** $G$ slightly increases when $s$ increases from 5 to 40, meaning $s$ = 5 is good enough.

The dashed line for $R_1(G)$ in **Figure 6** indicates that if we replace $P(x|a_j) = P(x|\theta_j, s)$ with a normal distribution, $P(x|\beta_j, s)$, $G$ and $G/R_1$ do not change significantly.

## 5. Discussion

### 5.1. Relationships between SVB and the Rate-Distortion Theory and the Maximum Entropy Principle

The MID iteration of SVB for $P(y|x)$ and $P(y)$ directly comes from the parameter solution of the rate-fidelity function $R(G)$ (see (28) and (29)). This iteration originated from Shannon et al.'s research on the information rate-distortion function $R(D)$. Although the constraint for $R(D)$ is only the distortion function, the constraint for $R(G)$ may be various learning functions, including likelihood, truth, membership, similarity, distortion, and copula density functions. We use these functions to construct the semantic information measure. Since $I(x; \theta_j) = \log[P(x|\theta_j)/P(x)]$, SVB is compatible with the maximum likelihood criterion.

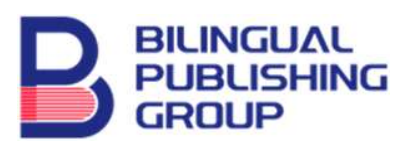

Given $P(x)$, minimizing Shannon's mutual information is equivalent to maximizing the posterior entropy $H(X|Y)$. Thus, the maximum information efficiency criterion used for SVB includes the maximum entropy criterion. Additionally, SVB allows the use of $s$ (see (28)) to strengthen constraints, which means it permits reducing $G/R$ while increasing $G$. Therefore, it can help us make the trade-off between information-efficiency maximization and semantic information (or purposiveness) maximization.

Compared to beta-VAE (where the objective function is $F_1 - \beta R$), SVB achieves information efficiency with an upper bound of 1, making it easier to balance maximizing semantic information with maximizing information efficiency.

### 5.2. Reasons for the Convergence of Mixture Models

Some researchers believe that the EM algorithm converges as the complete data log-likelihood $Q = -H(X, Y|\theta)$ increases [36,37]. Some researchers believe that continuously decreasing the free energy $F$ can cause mixture models to converge. However, the example in Section 4.1 reveals that the above conclusion is correct when $F$ is interpreted as $F_2$, but incorrect when $F$ is interpreted as $F_1$. The examples in Figure 3 show that $Q$ may decrease and $F_1$ may increase as the mixture model converges.

Section 3.2 provides a new convergence proof of mixture models. It affirms that decreasing $R - G$ (i.e., increasing the information efficiency $G/R$) makes the relative entropy $H(P||P_\theta)$ approach zero.

### 5.3. Comparison between VB and SVB

The primary tasks of VB and SVB are the same—using variational methods to solve latent variables based on observed data and constraints. The differences are:

1) VB uses the minimum free energy criterion—it uses either $P(y)$ or $P(y|x)$ as a variation, whereas SVB alternates between using $P(y|x)$ and $P(y)$ as variations.

2) SVB allows us to use $s$ to enhance constraints, which makes it easier to balance between maximizing semantic information ($G$) and maximizing information efficiency ($G/R$).

3) For SVB, the constraint functions are more diverse, including not only likelihood functions, but also truth, similarity, distortion, and copula density functions.

4) VB is based on Bayesian Inference (BI), considering the probabilities of model parameters, and uses mean-field approximation [38] to solve $P(y|x)$ and $P(y)$; whereas SVB does not consider the probabilities of parameters and uses the MID iteration to solve $P(y|x)$ and $P(y)$. The latter algorithm is simpler, but may not be suitable in some situations.

Because SVB is clearly compatible with the maximum likelihood criterion and the maximum entropy principle, it can be understood more easily.

### 5.4. Further Research Needed to Integrate SVB with Neural Networks

Although SVB has not yet been applied to neural networks, it has potential applications because it facilitates the use of similarity and truth functions. Interpreting neural network weights as probabilities can be challenging due to the normalization requirements of probability distributions. For instance, if we use a group of $m \times n$ weights to represent $n$ transition probability functions $P(y_j|x)$ or $P(\theta_j|x)$ ($j = 1, 2, \ldots$) with certain distributions, constructing such functions is difficult because of the normalization condition $\sum_j P(y_j|x)=1$ for every $x$. However, there is no normalization requirement when using weights to represent truth or similarity functions. All we need are the relative weights in each group. A group of weights can equal an exponential or negative exponential function.

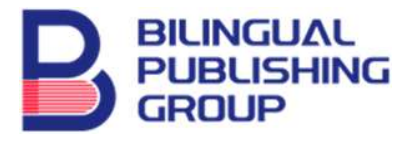

SVB employs the iterative method without requiring gradients. Theoretically, during the training phase, we can consider using a Gaussian channel mixture model to emulate a Restricted Boltzmann Machine (RBM), thereby simplifying computations. During the classification phase, we let $s\rightarrow\infty$ to obtain $P(y_j|x)$ ($j = 1, 2, \ldots, n$), which becomes a classification function. The practical effectiveness of this approach needs to be tested.

There are already many effective deep learning methods. Combining SVB with these methods is expected to achieve better results.

Beta-VAE [10] has achieved excellent results in deep learning. In contrast, SVB offers more explicit information efficiency. Combining SVB with deep learning methods holds promise for even better results.

## 6. Conclusions

Semantic Variational Bayes was derived from the parameter's solution of the rate-fidelity function $R(G)$. Its variational and iterative methods originated from Shannon and others' research on the rate-distortion function. Like Variational Bayes, SVB is used to solve the probability distribution of latent variables for given observed data and specific constraints. However, SVB allows various learning functions, including likelihood, truth, membership, similarity, and distortion, and copula density functions, to be used as constraints.

SVB and VB use similar optimization criteria (including the maximum semantic information criterion and the maximum information efficiency criterion). Still, SVB makes it easier to make the trade-off between maximum semantic information and maximum information efficiency.

The above conclusions were supported by experimental data presented in Section 4.

However, SVB does not consider the probability of model parameters, and it does not require exponential and logarithmic operations to optimize the Shannon channel. Both have their own unique advantages.

Further research and experimentation are needed to explore how SVB is applied to deep learning.

### Funding

This research received no external funding.

### Institutional Review Board Statement

Not applicable.

### Informed Consent Statement

Not applicable.

### Data Availability Statement

The data used in this study are available from the corresponding author upon reasonable request.

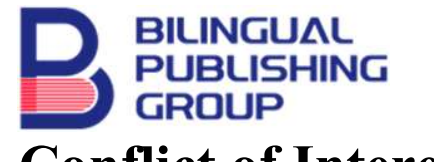

## Conflict of Interest

The Author declares no conflict of interest.

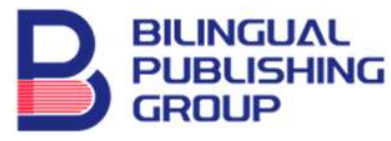